\documentclass[a4paper]{jpconf}

\usepackage{citesort}
\usepackage{graphicx}              
\usepackage{amsmath}               
\usepackage{amsfonts}              
\usepackage{amsthm}                
\usepackage{url}

\begin{document}

\nocite{*}

\title{Predicting the gender of Indonesian names}

\author{Ali Akbar Septiandri}

\address{Titik Koma Digital, Jakarta, Indonesia}

\ead{ali@titikkoma.tech}

\begin{abstract}
  We investigated a way to predict the gender of a name using character-level Long-Short Term Memory (char-LSTM). We compared our method with some conventional machine learning methods, namely Na\" ive Bayes, logistic regression, and XGBoost with n-grams as the features. We evaluated the models on a dataset consisting of the names of Indonesian people. It is not common to use a family name as the surname in Indonesian culture, except in some ethnicities. Therefore, we inferred the gender from both full names and first names. The results show that we can achieve 92.25\% accuracy from full names, while using first names only yields 90.65\% accuracy. These results are better than the ones from applying the classical machine learning algorithms to n-grams.
\end{abstract}

\section{Introduction}
Inspired by Liu and Ruths \cite{liu2013s}, we introduced a character-based approach to inferring one's gender from their name. In this study, we focused on Indonesian names. One of the main distinction between Indonesian names and English names is that it is not common to use family names as the surname. Therefore, we compared the two approaches, using only first names and full names, aspiring to get a better view to this problem.

In Indonesian names, we also face challenges like unisex names. For instance, ``Dwi'', ``Tri'', or ``Rizki'' and their variations can be used by either males or females. On the other hand, there are also some unmistakable names, such as ``Putra'' for males and ``Putri'' for females. Since Indonesian names come from several different roots, e.g. Sanskrit, Arabic, English, and Chinese, we might encounter a combination of names with those origins. Having said that, some people in Indonesia only have forename without a surname.

In this paper, we employed char-LSTM, a neural network architecture that process a sequence of characters, as a novel approach to tackle this classification problem and compared the result to the ones we got from conventional machine learning algorithms, such as logistic regression, Na\" ive Bayes, and XGBoost. We applied those algorithms to our self-labeled dataset. We focused only on gender classification in this study aspiring to get the best model to be applicable to cases where it can be useful to infer somebody's gender from his/her name. Examples include better customer interaction and estimating gender-related demographics in social media.

\section{Related Work} 
\label{sec:related_work}

This work is mainly based on \cite{liu2013s}. Although there are many gender classification studies driven by the emergence of social media like Twitter, this prediction task with names as one of the attributes is not common. Some of the work that is related to this task is \cite{burger2011discriminating}, \cite{alowibdi2013empirical}, and \cite{bergsma2013broadly}.

Liu and Ruths \cite{liu2013s} built a gender-name association score from first names in their study and stated that this is ``a promising approach since the measure compactly encapsulates a prior
on the gender of the individual based on their name.'' In \cite{alowibdi2013empirical}, first names are proposed as one of the attributes, but they convert the names to phoneme sequences and then generate n-grams of the phoneme sequences. On the other hand, Burger et al. chose to create 1-5 character n-grams and word unigram from full names \cite{burger2011discriminating}. Bergsma et al. also tried binary 1-4 character n-grams and used first names and last names binary token \cite{bergsma2013broadly}.

We believed that the success of LSTM \cite{hochreiter1997long} in several complex cases, such as shown in \cite{sutskever2011generating} and \cite{graves2013generating}, and neural networks in general could be beneficial in this rather simple study. By doing this, there is no need to extract n-grams from the names. We also explored character-level embedding inspired by \cite{kim2015character} yearning for better results. The idea of the char-LSTM architecture is based on the IMDB sentiment classification task provided in \cite{chollet2015keras}.


\section{Methodology} 
\label{sec:methodology}

\subsection{Dataset} 
\label{sub:dataset}

The names and associated gender data used in this study are compiled from graduate tracer studies. The resulting dataset consists of 4580 males (66.56\%) and 2301 females (33.44\%), for a total of 6881 names. The dataset is not released as part of this study as there is an obvious privacy concern regarding identifiable names and gender.

We acknowledge that there are certainly biases inherent in the data collection process, e.g. gender proportion and ``style'' of the names themselves do not reflect Indonesia's general population. However, the methodology of this study can still be applied and further research using more representative dataset is recommended.


\subsection{Features} 
\label{sub:features}

First of all, we removed some symbols, e.g. apostrophe ('), period (.), and hyphen (-), in the names. We also set the names to be written in lowercase characters. This is because it is very rare to find Indonesian names with mixed case.

Aside from the 2-5 character n-gram features as shown in prior work, we proposed simpler features, such as the first and last characters of first and last names that we call \emph{basic} features. We will see later in this study that by using these basic features, we can already achieve fair results. For these features, we transformed them using one-hot-encoder before using them as the classifiers' attributes.

For the 2-5 n-grams, we chose only 1000 top features with the highest values of chi-squared statistic test. This was done to remove n-grams ``that are most likely to be independent of class'' \cite{pedregosa2011scikit}. Also, this can make the learning process faster.

\begin{figure}[htbp]
\begin{center}
\includegraphics[width=\columnwidth]{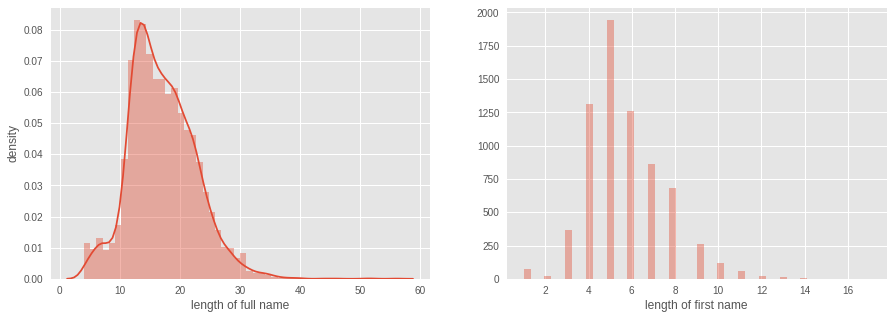}
\caption{Length of the names}\label{fig:length}
\end{center}
\end{figure}

For the char-LSTM classifier, we extracted the index for every character in the names. This will yield lists of characters with different length. Therefore, we padded the lists with zeros until they have the same length as the maximum number of characters in the dataset, which in our case is 56. For first names only, the maximum number of characters is 17. The distribution of the names by the number of characters can be seen in Figure~\ref{fig:length}.

\subsection{Algorithms} 
\label{sub:algorithms}

We employed neural networks implementation known as Keras \cite{chollet2015keras} in our experiments. Using embedding layer and LSTM, we passed the inputs and then use simple logistic regression as the output layer. We trained this model with Adam \cite{kingma2014adam} as the optimisation routine using default parameters and mini-batch size 32. We also tried different number of hidden units for the embedding layer and LSTM. For each layer, we tried 64, 128, and 256 for full names, and 32, 64, and 128 for first names.

On the other hand, we used scikit-learn \cite{pedregosa2011scikit} for the conventional machine learning algorithms other than XGBoost. We chose Na\" ive Bayes for its simplicity (almost no hyperparameter tuning needed) \cite{rish2001empirical} and logistic regression as the ``simpler version'' of neural networks. XGBoost \cite{Chen2016}, however, was chosen because of its decent performance in lots of machine learning competitions. For the conventional machine learning algorithms, we tuned the hyperparameters using options as shown in Table~\ref{tab:hyperparameters}. This was done by grid search and cross-validation.

\begin{table}[hptb]
\caption{List of hyperparameters}
\label{tab:hyperparameters}
\centering
\begin{tabular}{l l l}
  \br
  Classifier & Hyperparameter & Values tested \\
  \mr
  LogReg & \texttt{penalty} & ['l1', 'l2'] \\
         & \texttt{C} & [0.01, 0.1, 1, 10, 100] \\
  XGBoost & \texttt{max\_depth} & [3, 4, 5, 6, 7, 8, 9, 10] \\
          & \texttt{min\_child\_weight} & [0, 0.1, 1, 100,1000] \\
          & \texttt{gamma} & [0, 0.1, 1, 100,1000] \\
  \br
\end{tabular}
\end{table}


\subsection{Evaluation Approach} 
\label{sub:evaluation_approach}

Since we found this dataset to be unbalanced, aside from accuracy, we calculated the $F_1$-score as well. $F_1$-score is defined as:
$$
F_1 = 2 \cdot \frac{\text{precision} \cdot \text{recall}}{\text{precision} + \text{recall}}
$$
We tried the algorithms on features from full names. Our goal is to produce the best model possible from these features. Then, we attempted to use first names only as the features to see whether we would be able to achieve similar results. In the first experiment with LSTM, we set the number of the epoch to be 20.



\section{Results} 
\label{sec:results}

\subsection{Inferring Gender from Full Names} 
\label{sub:inferring_gender_from_full_names}

Using the tuned algorithms with basic features and n-grams, we got the results as shown in Table~\ref{tab:conv_full}. We can see that we achieved the best performance for Na\" ive Bayes and logistic regression using 3-gram features, while using 2-gram shows a slightly better result for XGBoost.

\begin{table}[hptb]
\caption{Results from basic and n-gram features}
\label{tab:conv_full}
\centering
\begin{tabular}{l c c c c c c}
  \br
           & \multicolumn{3}{c}{Accuracy (\%)} & \multicolumn{3}{c}{$F_1$ (\%)} \\
  Features & NB & LogReg & XGBoost & NB & LogReg & XGBoost \\
  \mr
   basic & 78.69 & 80.44 & 81.50 & 84.07 & 85.59 & 86.35 \\
  2-gram & 79.81 & 84.21 & 83.87 & 85.10 & 88.63 & 88.55 \\
  3-gram & 84.12 & \textbf{85.28} & 83.39 & 88.35 & \textbf{89.44} & 88.30 \\
  4-gram & 83.78 & 83.54 & 79.90 & 88.52 & 88.33 & 86.57 \\
  5-gram & 80.77 & 80.92 & 76.71 & 86.93 & 86.96 & 85.22 \\
  \br
\end{tabular}
\end{table}

\begin{figure}[htp]
\begin{center}
\includegraphics[width=\columnwidth]{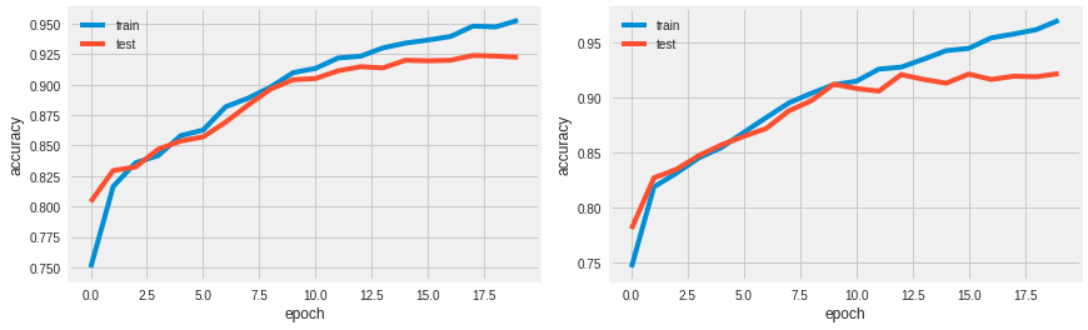}
\caption{The effect of epoch to accuracy for the best (left) and second best (right) models}\label{fig:epoch}
\end{center}
\end{figure}

In the next step, we trained the char-LSTM model using different number of hidden units. By doing this, we got two best results with slightly different perforamance. The best one we got is from using $\mathbb{R}^{256}$ vector embedding and 64 hidden units for LSTM, while the second best is from vector embedding in $\mathbb{R}^{128}$ and 256 hidden units for LSTM. The results as well as the effect of the number of epochs to the accuracy can be seen in Figure~\ref{fig:epoch}. The accuracy and $F_1$ for the best model are \textbf{92.25\%} and \textbf{94.35\%} respectively, while the second best model yields 92.20\% accuracy and 94.31\% $F_1$.

\subsection{Inferring Gender from First Names Only} 
\label{sub:inferring_gender_from_first_names_only}

Similar trends can be seen in the results from features extracted from first names. As we can see in Table~\ref{tab:conv_first}, the 3-gram feature outperforms other features when using Na\" ive Bayes and logistic regression. On the XGBoost case, 2-gram yields better result than 3-gram which is similar to the result from using full names.

\begin{table}[hptb]
\caption{Results from basic and n-gram features of first names}
\label{tab:conv_first}
\centering
\begin{tabular}{l c c c c c c}
  \br
           & \multicolumn{3}{c}{Accuracy (\%)} & \multicolumn{3}{c}{$F_1$ (\%)} \\
  Features & NB & LogReg & XGBoost & NB & LogReg & XGBoost \\
  \mr
   basic & 74.77 & 75.30 & 76.85 & 81.01 & 81.73 & 83.31 \\
  2-gram & 78.64 & 81.60 & \textbf{82.37} & 84.51 & 87.27 & \textbf{87.80} \\
  3-gram & 79.85 & 81.50 & 78.35 & 85.95 & 87.17 & 85.85 \\
  4-gram & 77.34 & 77.38 & 72.74 & 85.06 & 85.18 & 83.06 \\
  5-gram & 72.59 & 72.69 & 69.93 & 82.93 & 83.02 & 81.80 \\
  \br
\end{tabular}
\end{table}

\begin{figure}[htp]
\begin{center}
\includegraphics[width=.5\columnwidth]{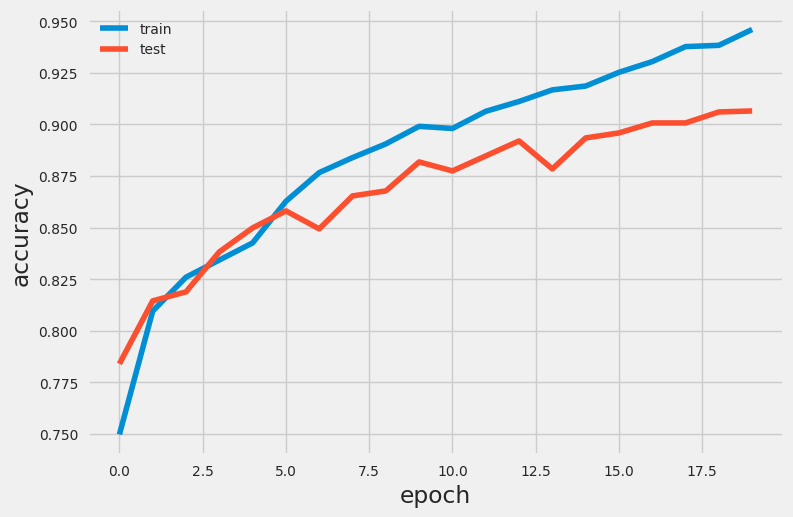}
\caption{The effect of epoch to accuracy for the best model using char-LSTM}\label{fig:nn}
\end{center}
\end{figure}

The results with char-LSTM on first names are also promising. The best model in this experiment was from using $\mathbb{R}^{128}$ vector embedding and 128 hidden units for LSTM. With this architecture, we reached \textbf{90.65\%} accuracy and \textbf{93.19\%} $F_1$. This means that the $F_1$ is about 1.16\% off from the best model produced with full names. The effect of epoch in the best model from first names can be seen in Figure~\ref{fig:nn}.

\subsection{Exploration on Specific Names} 
\label{sub:exploration_on_specific_names}

Since we have seen that the best model came from the char-LSTM algorithm, we wanted to know how this model performs in several cases. One of the advantages of using the char-LSTM is that you can see the changes in the class probability for every character addition. Therefore, we fed the best model trained with full names to predict some of the names. Bear in mind that these names can be in the first, middle, or last part of the names.

The model we used in this experiment is the one with 256 and 64 hidden units for embedding and LSTM layers respectively. Probabilities produced from this model show the likelihood of whether the name is a male. Therefore, $$P(y=\text{female}|\text{name}) = 1 - P(y=\text{male}|\text{name})$$

\begin{figure}[htbp]
\begin{center}
\includegraphics[width=.5\columnwidth]{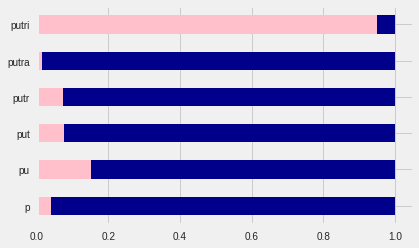}
\caption{Char-LSTM evaluation on specific names}\label{fig:incremental}
\end{center}
\end{figure}

We evaluated the name `Putra' and `Putri'. As mentioned before, these names are almost unambiguous, i.e. `Putra' is for males and `Putri' is for females. In the dataset, there are 215 people with `putra' in their name and 100 females have `putri' in their name. Only one name that contains `Saputra' and is a female. The evaluation per one character increment can be seen in Figure~\ref{fig:incremental}. The blue and pink bars represent the probability of the name being male or female respectively. 

\section{Conclusions and Future Work} 
\label{sec:conclusions_and_future_work}

Our work has shown that using char-LSTM can improve the model for predicting gender from names compared to only using conventional machine learning algorithms with n-grams. We managed to increase the best accuracy from 3-gram using logistic regression, i.e. 85.28\%, to 92.25\% using char-LSTM. This means that we reduced the remaining error rate by 47.35\%. The difference between the best $F_1$-score from n-grams and char-LSTM in this study is 4.91\% (from 89.44\% to 94.35\%).

In the second experiment, we found out that using first names only might be enough. With even fewer features, the char-LSTM could produce 90.65\% accuracy. This level of accuracy also outperformed the results from n-grams. However, the first experiment suggested that using full names is better if possible, i.e. when the surnames could be gender-specific.

There are several ways to improve this study, e.g. using different nationalities of the names or using usernames from social media. We would also like to see how this approach could improve gender prediction in general given other features from social media. Furthermore, we believe that these methods can be extended to predicting other things, such as the ethnicity of the people. It would also be interesting to build a generative model, e.g. using conditional Generative Adversarial Networks, to generate names given the gender or the ethnicity.

\section*{Acknowledgement}

The author would like to thank Okiriza Wibisono and Yosef Ardhito Winatmoko for their advice and support to this study.

\section*{References}

\bibliographystyle{iopart-num}

\bibliography{gender}

\end{document}